\documentclass[conference]{IEEEtran}
\IEEEoverridecommandlockouts
\usepackage{cite}
\usepackage{amsmath,amssymb,amsfonts}
\usepackage{algorithmic}
\usepackage{graphicx}
\usepackage{textcomp}
\usepackage{xcolor}
\usepackage{multicol,multirow,booktabs,mathabx}
\def\BibTeX{{\rm B\kern-.05em{\sc i\kern-.025em b}\kern-.08em
    T\kern-.1667em\lower.7ex\hbox{E}\kern-.125emX}}
\begin{document}

\title{Adaptive Confidence Threshold for ByteTrack in Multi-Object Tracking
}

\author{\IEEEauthorblockN{Linh Van Ma\IEEEauthorrefmark{1}, Muhammad Ishfaq Hussain\IEEEauthorrefmark{1}, JongHyun Park\IEEEauthorrefmark{1}, Jeongbae Kim\IEEEauthorrefmark{2}, Moongu Jeon\IEEEauthorrefmark{1}}
\IEEEauthorblockA{\IEEEauthorrefmark{1}\textit{School of Electrical Engineering and Computer Science, Gwangju Institute of Science and Technology, Republic of Korea}\\
\IEEEauthorrefmark{2}\textit{Department of Mechanical-IT Convergence System Engineering, Pusan National University, Busan, Republic of Korea}\\
\IEEEauthorrefmark{1}\{linh.mavan, ishfaqhussain, citizen135, mgjeon\}@gist.ac.kr, \IEEEauthorrefmark{2}j0k9262@gmail.com}}

% \author{\IEEEauthorblockN{Linh Van Ma, Muhammad Ishfaq Hussain, JongHyun Park, Moongu Jeon}
% \IEEEauthorblockA{\textit{School of Electrical Engineering and Computer Science}, \\
% \textit{Gwangju Institute of Science and Technology, Republic of Korea}\\
% \{linh.mavan, ishfaqhussain, citizen135, nmgjeon\}@gist.ac.kr}
% }

\maketitle

\begin{abstract}
We investigate the application of ByteTrack in the realm of multiple object tracking. ByteTrack, a simple tracking algorithm, enables the simultaneous tracking of multiple objects by strategically incorporating detections with a low confidence threshold. Conventionally, objects are initially associated with high confidence threshold detections. When the association between objects and detections becomes ambiguous, ByteTrack extends the association to lower confidence threshold detections. One notable drawback of the existing ByteTrack approach is its reliance on a fixed threshold to differentiate between high and low-confidence detections. In response to this limitation, we introduce a novel and adaptive approach. Our proposed method entails a dynamic adjustment of the confidence threshold, leveraging insights derived from overall detections. Through experimentation, we demonstrate the effectiveness of our adaptive confidence threshold technique while maintaining running time compared to ByteTrack.
\end{abstract}


\section{Introduction}
Visual multi-object tracking is a challenging task. Many filtering tracking \cite{reid1979algorithm,fortmann1983sonar,vo2016efficient} algorithms have been well developed and can track multiple targets in noisy environments such as military radars and satellites. However, filtering techniques applied in visual tracking \cite{kim2019labeled,abbaspour2022online} are much more complicated compared to recent simple developed algorithms for visual tracking such as SORT \cite{wojke2017simple}. This simple algorithm employs a Kalman filter to predict and update object motion. An association algorithm such as the Hungarian algorithm \cite{kuhn1955hungarian} (or LapJV \cite{jonker1988shortest}) is used to associate measurements with tracks. This simple approach is effective because recent visual multi-Object detectors \cite{zhang2021fairmot,ge2021yolox,redmon2016you,wang2020towards,zhang2022bytetrack} are highly accurate with few false detections.

In this paper, we follow the simple approach with the state-of-the-art tracker, ByteTrack \cite{zhang2022bytetrack}. It is filtering-based approaches, the primary branch of motion model-centered tracking. It relies on a predictive transition function for anticipating object states in upcoming time steps, known as "estimation". It additionally utilizes an observation model, like an object detector, to obtain measurements of object states known as "observation". These observations play a crucial role in refining the posterior parameters of the filter. We recognize that ByteTrack uses different confidence scores for different sequences of the MOT Challenge dataset \cite{milan2016mot16}. They have not specified how to obtain confidence thresholds. A naive approach is that we repeatedly run ByteTrack with different confidence thresholds and then get the threshold with the highest tracking performance. However, we can only tune the confidence score threshold if we have prior knowledge of a video. This limitation can degrade the performance of ByteTrack in real-world applications. Hence, we propose a method to overcome this limitation by using an adaptive confidence threshold. In computer vision, similar recent works  \cite{el2022multiple,stalder2010cascaded} proposed a dynamic confidence cut-off score method in implementing ByteTrack. Given an input detector geometric filter, the authors successively filter the detection confidence responses by including spatial and temporal context using a geometric filter, background filter, trajectory filter, and post-processing. Subsequently, the refined detection confidence can be used to track multiple objects. Their approach is complicated compared to the simple yet effective method of ByteTrack.

\begin{figure*}%[htbp]
\centerline{\includegraphics[width=\textwidth]{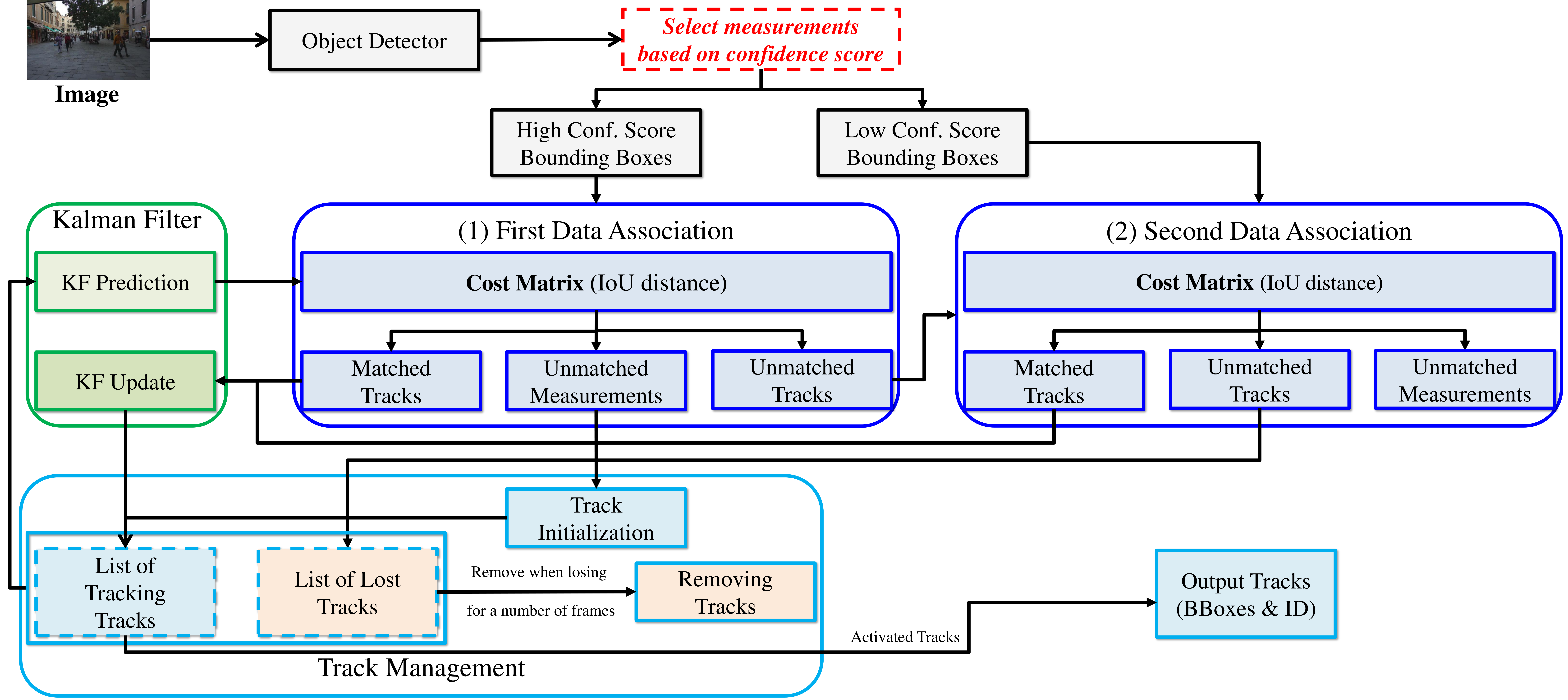}}
\caption{An overview of ByteTrack. It extends the SORT algorithm \cite{bewley2016simple} with two lists of measurements (bounding boxes with low and high confidence scores). It has two data associations. The first one associates tracking objects with a high confidence score. The second one associates unmatched objects (from the first association) with a low confidence score. Our proposed method is highlighted in the dash-red rectangle box.}
\label{fig:bytetrack}
\end{figure*}

Our contributions are summarized as the following:
\begin{enumerate}
    \item We recognize, analytically and empirically, a limitation of ByteTrack that uses a fixed confidence threshold to classify low and high confidence detections;
    \item We propose an adaptive confidence threshold method to classify low and high-confidence detected measurements.
\end{enumerate}

The paper is organized as follows. In Section \ref{sec:backgrounds} related works in multi-object tracking are discussed. In Section \ref{sec:adaptiveconf}, we present our proposed adaptive confidence threshold method. Experiments on the challenge dataset to verify the effectiveness of our proposed method are given in Section \ref{sec:experiment}, and Section \ref{sec:conclusion} concludes the paper. 

\section{Related Works}\label{sec:backgrounds}
\textbf{Motion Models based Multi-object Tracking}.
SORT \cite{bewley2016simple} proposes a simple method that only employs Kalman Filter \cite{kalman1960contributions} to predict the motion of objects and the Hungarian algorithm \cite{kuhn1955hungarian} to solve tracks to measurements assignment. The simplicity of this tracking method enables a rapid running time of 260 Hz, surpassing the speed of other leading trackers by more than 20 times. However, SORT only considers good detections (bounding boxes with a confidence score greater than a threshold) and can potentially ignore occluded objects with a low confidence score. Hence, ByteTrack \cite{zhang2022bytetrack} further improves SORT by considering associating detected bounding boxes with lower confidence scores.

IOU Tracker\cite{bochinski2017high} takes a different approach similar to global nearest neighbor (GNN) \cite{blackman1999design}. It associates a tracking object with the nearest detection (using IoU, Intersection Over Union). V-IOU tracker further extends IoU tracker by employing a visual tracker to compensate for missing object locations \cite{bochinski2018extending}. Their proposed algorithm showcases the ability to achieve remarkable speeds, reaching up to 100,000 frames per second, while also surpassing the performance of state-of-the-art approaches on the DETRAC vehicle tracking dataset \cite{lyu2017ua}.

\textbf{Re-Identification Feature-based Multi-object Tracking}.
DeepSORT \cite{wojke2017simple} introduces enhancements to SORT by incorporating appearance information, which results in improved tracking performance. This extension allows for better tracking of objects during occlusions, leading to a reduction in the frequency of identity switches. Experimental evaluation demonstrates that these extensions lead to a significant reduction (45\%) in identity switches, while still achieving competitive overall performance, especially at high frame rates. MOTDT \cite{chen2018real}, FairMOT \cite{zhang2021fairmot}, JDE \cite{wang2020towards}, GSDT \cite{wang2021joint} follows the same mechanism of DeepSORT. 

\textbf{Filtering-based Visual Multi-object Tracking}. MHT \cite{reid1979algorithm,kim2015multiple} considers more than one measurement (within the validation gate of a track) as new track hypotheses. However, this leads to an unbounded solution set because the framework does not support track management. In JPDA \cite{fortmann1983sonar,musicki2004joint,bar2005tracking}, each track (hypothesis) is maintained based on a quality measure (object existence probability) and therefore results in a dynamically changing but limited number of tracks. Using the Labeled Random Finite Set Statistic (LFRS) approach, the authors \cite{kim2019labeled} propose a filter that updates tracks with detections but switches to image data when detection loss occurs, thereby exploiting the efficiency of detection data and the accuracy of image data. The authors \cite{abbaspour2022online} employ \cite{vo2016efficient} $\delta$-GLMB to propose an occlusion handling module resulting lower identity switch compared to the stat-of-the-art method. Although these filtering methods can handle complex scenarios (such as a high number of clutters), their runtime is much slower without a big performance improvement gap compared to the state-of-the-art simple MOT tracking algorithm such as SORT \cite{bewley2016simple}. The reason is that video data has a high frame rate. Hence, we only need one data association between tracks and measurements. 

\section{Adaptive Confidence Threshold for ByteTrack}\label{sec:adaptiveconf}
In this section, we review ByteTrack, discuss how the confidence score is calculated in YOLOX \cite{ge2021yolox}, and subsequently present our proposal method to calculate the adaptive confidence score threshold. 

An overview of the ByteTrack tracking algorithm is shown in Fig. \ref{fig:bytetrack}. Similar to SORT, it assumes that measurements closely appear near a tracking object. Hence, it employs IoU to compute the distance between tracks and measurements. It has two data association rounds (shown Fig. \ref{fig:bytetrack}, two blue blocks). The first one associates current tracking objects with a high confidence score. It extends the SORT algorithm \cite{bewley2016simple} by adding a second data association for tracks with lower confidence score detections. Specifically, the second association tries to connect unmatched objects (from the first association) with a low confidence score. If no unmatched tracks from the second data association, they mark them as lost tracks. It will be kept for a certain number of frames in a separate memory (such as a list). These lost tracks can be matched with upcoming measurements. If no matched pair between tracks and measurements happens during the keeping period, it will be discarded.

In ByteTrack, the authors classify high and low-detected bounding boxes by a predefined threshold. This threshold can be determined by running repeated experiments and obtaining the one with the highest tracking accuracy. This method can limit tracking performance when a scene changes (e.g. an experiment is done in a video captured from rare places but changed to dense and crowned places). We observe that they use different thresholds for different sequences in the same dataset (e.g. MOT16, MOT16-02, MOT16-09). More specifically, the authors set the confidence threshold to 0.6 for the MOT16 training dataset while setting a lower value to 0.3 for MOT20. The MOT20 dataset is denser and objects are heavier occluded than MOT16. Hence, the YOLOX detector can have higher confidence in detecting objects in MOT16 than in MOT20. Since we already know how crowded the MOT20 dataset is, we can guess and tune the confidence threshold correctly. However, in real applications, we generally have no knowledge of a scene leading to a hard decision to select the confidence threshold. Hence, we need a method that adaptively classifies low and high-confidence score detections outputted from deep learning detectors.

In YOLOX \cite{ge2021yolox} (an anchor-free deep learning detector based on YOLO\cite{redmon2016you}), at test time, the authors multiply the conditional class probabilities and the individual box,
\begin{equation}
    P(C_i|O) * Pr(O) * IOU^{truth}_{pred} = P(C_i) * IOU^{truth}_{pred}
\end{equation}

which gives them class-specific confidence scores for each box. These scores encode both the probability of that class appearing in the box and how well the predicted box fits the object. Hence, for truth-positive detections, $P(C_i)$ and $IOU^{truth}_{pred}$ have high values resulting a high confidence score. In contrast, for false positive detections, these values are indeed small since the IOU (Intersect Over Union) overlap between a false target and its ground truth is relatively small because the most appropriate bounding boxes are filtered by the Non-Maximum Suppression algorithm. Hence, there may be a good separation between true positive and false positive detections.

\begin{figure}%[htbp]
\centerline{\includegraphics[width=0.45\textwidth]{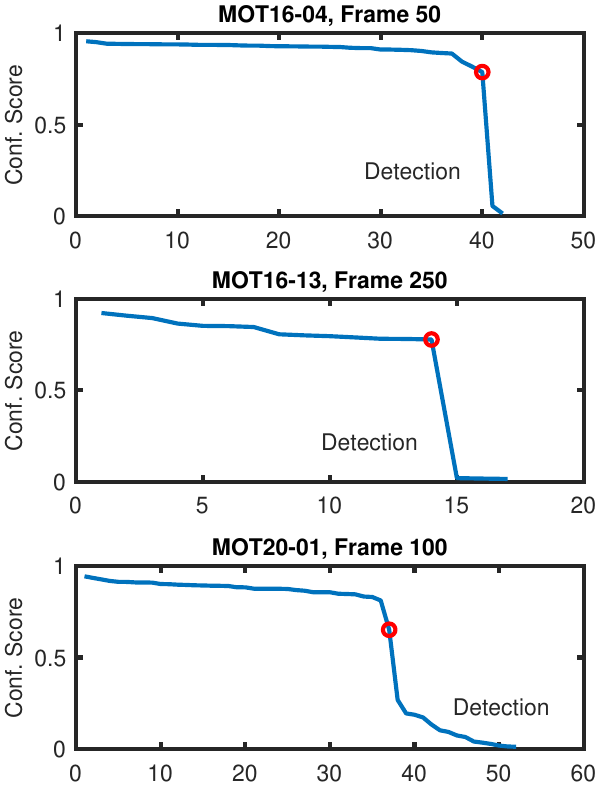}}
\caption{Example of our adaptive confidence threshold. The red circle specifies that detections with a confidence score higher than that point are selected as high and the rest are detections with low confidence scores.}
\label{fig:adapconf}
\end{figure}

By examining the confidence threshold (outputted from YOLOX), we visually notice the separation between high and low detected bounding boxes can be logically calculated as shown in Fig. \ref{fig:adapconf}. Given a list of confidence scores corresponding to detected objects, we detect the steepest drop from the high to the low confidence score. In each frame, we classify the low and high-confidence bounding boxes by finding the steepest point \cite{meza2010steepest}. %Hence, it can adapt to any changes in scenes or datasets.

In our proposal adaptive confidence score, given $c_i, (i\in \mathbb{N})$ is the confidence score of detection $i$, we first sort these values in decreasing order and obtain $c^s=\{c^s_i\}$. Suppose $f(c^s_i)$ is the function that represents the confidence score curve. If $c^s_i \leq c^s_j$, then $f(c^s_i)\leq f(c^s_j)$. We employ the steepest descent method to find the point that separates the confidence score into high and low. Hence, the confidence threshold can be found by taking the derivative of the confidence threshold function $f(.)$. We approximate this process by calculating the $1^{st}$ discrete difference of $c^s_i$. The minimum difference specifies the confidence threshold where the steepest point is located. Hence, the adaptive threshold is $c_{i_{th}}$, where $i_{th}= \underset{j}{\mathrm{argmin}} (c^s_{j+1} - c^s_{j})$.

\begin{figure*}%[htbp]
\centerline{\includegraphics[width=\textwidth]{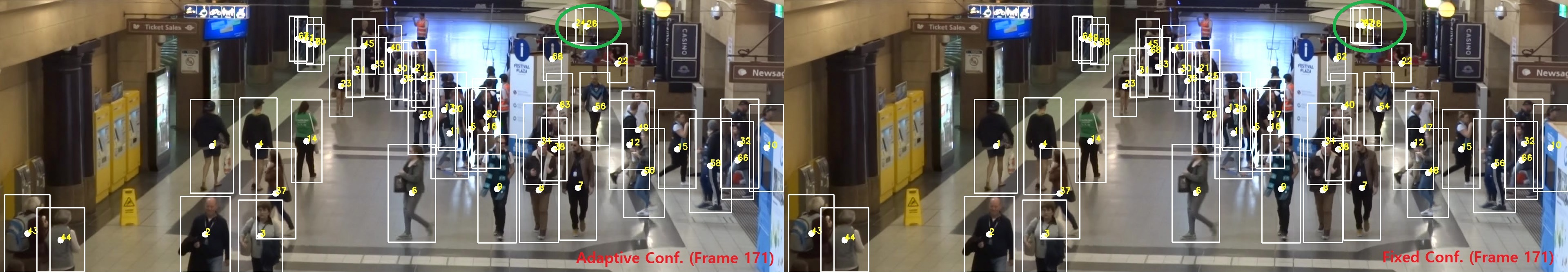}}
\caption{Visualization of the difference between adaptive and fixed confidence in MOT20-01 sequence dataset (Frame number 171). We highlight the difference in a green ellipse (best view in zoom). The fixed confidence method produces two more false tracks compared to ours.}
\label{fig:visualcompare}
\end{figure*}

\begin{figure}%[htbp]
\centerline{\includegraphics[width=0.5\textwidth]{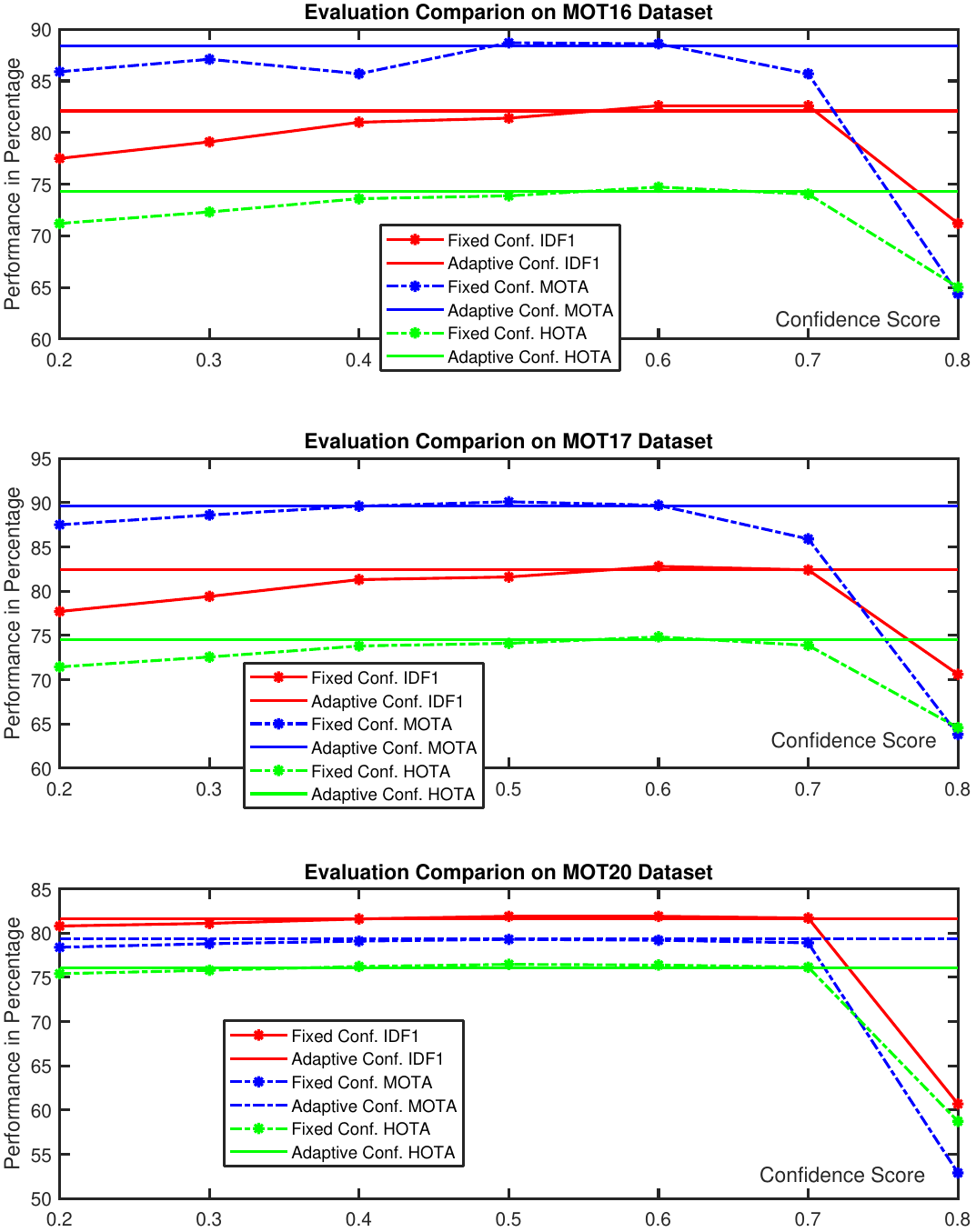}}
\caption{Experiments comparison on three datasets (MOT16, MOT17 and MOT20). The results show that our method does not need to tune the confidence score but achieves tracking performance as high as the best result of ByteTrack with confidence score tunning.}
\label{fig:expconf}
\end{figure}

\begin{table*}
\centering
\caption{Evaluation results on the test set of MOT16, MOT17 and MOT20. Our results are similar to ByteTrack without tuning the confidence process. Note that ByteTrack sets confidence thresholds for different sequences.}
    \label{tbl:testset}
\begin{tabular}{l|l|c|c|c|c|c|c|c|c|c}
\hline
     Dataset & Method    & MOTA$\uparrow$ & IDF1$\uparrow$ & HOTA$\uparrow$ & MT$\uparrow$          & ML$\downarrow$        & FP$\downarrow$     & FN$\downarrow$     & ID SW $\downarrow$    & Frag $\downarrow$       \\
\hline
\multirow{2}{*}{MOT16}     &  ByteTrack      & 78.2 & 77.3 & 62.9 & 366(48.2)   & 124(16.3) & 9,308  & 29,722 & 800(0.0)   & 1,130(0.0)  \\
    &   Ours & 77.9 & 77.1 & 62.6 & 379(49.9)   & 118(15.5) & 10,681 & 28,659 & 904(0.0)   & 1,214(0.0)  \\
\hline
\multirow{2}{*}{MOT17}  &   ByteTrack    & 78.6 & 76.9 & 62.6 & 1,101(46.8) & 411(17.5) & 19,185 & 99,225 & 2,502(0.0) & 3,531(0.0)  \\
    &   Ours & 78.6 & 76.8 & 62.3 & 1,134(48.2) & 381(16.2) & 22,386 & 95,283 & 2,874(0.0) & 3,867(0.0) \\
\hline
\multirow{2}{*}{MOT20}  &   ByteTrack    & 76   &  75.5  & 	61.3  & 841(67.7)  & 121(9.7)	  & 25,161  & 97,524  & 1,443(0.0)  & 3,124(0.0)  \\
    &   Ours & 76 & 	75.4 & 	61.3 & 	841(67.7) & 	122(9.8) & 	24,954 & 	97,547 & 	1,437(0.0) & 	3,115(0.0) \\
\hline
\end{tabular}
\end{table*}

\section{Experiments}\label{sec:experiment}

\textbf{Datasets}. We evaluate our method on multiple multi-object tracking datasets including MOT16, MOT17 \cite{milan2016mot16}, and MOT20 \cite{dendorfer2020mot20}.

\textbf{Metrics}. We adopt three main tracking performance scores. First, the HOTA \cite{luiten2021hota} score maintains a proper balance between the accuracy of data association and object detection. Second, IDF1 \cite{ristani2016performance} represents how robust a tracker can maintain object identity throughout a number of frames. Finally, MOTA \cite{bernardin2008evaluating} is highly related to the detection performance of trackers.

In this section, we report the benchmark results on the three challenge datasets. To make a fair comparison, we use the same private detection as ByteTrack \cite{zhang2022bytetrack}, and YOLOX \cite{ge2021yolox}. The results show that our adaptive strategy can achieve performance comparable to ByteTrack without the need to fine-tune the confidence score. %ByteTrack obtains the best performance in terms of IDF1, MOTA, HOTA if the detection confidence score is set to 0.5 or 0.6. In MOT20, ByteTrack shows the best result for detection confidence score between 0.4 and 0.6. If the confidence score is higher than 0.7 the tracking performance steeply degrades.

Fig. \ref{fig:visualcompare} visually shows the comparison between our adaptive confidence method and fixed confidence threshold in the MOT20-01 sequence dataset. ByteTrack produces more false tracks since the confidence threshold is set to lower than ours. Fig. \ref{fig:expconf} illustrates that in MOT16 and MOT17, ByteTrack obtains the best performance when the confidence threshold is set in the range [0.5, 0.6]. However, the performance dramatically drops if we increase the confidence threshold to higher than 0.7. Our method does not need to set a threshold but considerably archives a high-performance result that ByteTrack can achieve. In MOT20, ByteTrack maintains its best performance between a confidence threshold of 0.3 to 0.5. Our adaptive strategy can achieve similar performance compared to ByteTrack.

We also test our method on MOT16, MOT17, and MOT20 testing datasets. For a fair comparison, we use source code provided by the ByteTrack authors \cite{zhang2022bytetrack} and strictly set up parameters following their instructions. All tracking results are submitted to the MOTChallenge website\cite{dendorfer2021motchallenge}. Tab. \ref{tbl:testset} shows that our performance results are similar (less than 0.3\% of performance differences) to ByteTrack without the need to tune the confidence process. Note that Bytetrack sets confidence threshold for different sequences. Specifically, in MOT16 dataset thresholds are set as \{MOT16-01: 0.65, MOT16-06: 0.65, MOT16-12': 0.7, MOT16-14: 0.67\} and the rest sequences are set to 0.6. In the MOT20 dataset, all sequences with detection threshold are set to 0.3. As our method is designed to be simple for better generalization, our implementation maintains running time compared to ByteTrack.

\section{Conclusion}\label{sec:conclusion}
In this paper, we review the popular tracking method SORT and the state-of-the-art ByteTrack. We recognize a ByteTrack limitation of using a fixed (fine-tuned) confidence score. We propose an adaptive confidence threshold method to overcome the limitation. Our result shows that we can achieve tracking performance as high as the fine-tuned strategy can obtain while maintaining its running time. In the future, this method can be extended to various detectors by examining how the detection confidence threshold is calculated.

\section*{Acknowledgements}
This work was supported by the Institute of Information \& Communications Technology Planning \& Evaluation (IITP) grant funded by the Korean government (MSIT) (No. 2014-3-00077). In addition, this work was supported by Culture, Sports and Tourism R\&D Program through the Korea Creative Content Agency grant funded by the Korea government (MCST) in 20xx (R2022060001).

\bibliographystyle{IEEEtran}
\bibliography{reflib}

% Generated by IEEEtran.bst, version: 1.14 (2015/08/26)
\begin{thebibliography}{10}
\providecommand{\url}[1]{#1}
\csname url@samestyle\endcsname
\providecommand{\newblock}{\relax}
\providecommand{\bibinfo}[2]{#2}
\providecommand{\BIBentrySTDinterwordspacing}{\spaceskip=0pt\relax}
\providecommand{\BIBentryALTinterwordstretchfactor}{4}
\providecommand{\BIBentryALTinterwordspacing}{\spaceskip=\fontdimen2\font plus
\BIBentryALTinterwordstretchfactor\fontdimen3\font minus
  \fontdimen4\font\relax}
\providecommand{\BIBforeignlanguage}[2]{{%
\expandafter\ifx\csname l@#1\endcsname\relax
\typeout{** WARNING: IEEEtran.bst: No hyphenation pattern has been}%
\typeout{** loaded for the language `#1'. Using the pattern for}%
\typeout{** the default language instead.}%
\else
\language=\csname l@#1\endcsname
\fi
#2}}
\providecommand{\BIBdecl}{\relax}
\BIBdecl

\bibitem{reid1979algorithm}
D.~Reid, ``An algorithm for tracking multiple targets,'' \emph{IEEE
  transactions on Automatic Control}, vol.~24, no.~6, pp. 843--854, 1979.

\bibitem{fortmann1983sonar}
T.~Fortmann, Y.~Bar-Shalom, and M.~Scheffe, ``Sonar tracking of multiple
  targets using joint probabilistic data association,'' \emph{IEEE journal of
  Oceanic Engineering}, vol.~8, no.~3, pp. 173--184, 1983.

\bibitem{vo2016efficient}
B.-N. Vo, B.-T. Vo, and H.~G. Hoang, ``An efficient implementation of the
  generalized labeled multi-bernoulli filter,'' \emph{IEEE Transactions on
  Signal Processing}, vol.~65, no.~8, pp. 1975--1987, 2016.

\bibitem{kim2019labeled}
D.~Y. Kim, B.-N. Vo, B.-T. Vo, and M.~Jeon, ``A labeled random finite set
  online multi-object tracker for video data,'' \emph{Pattern Recognition},
  vol.~90, pp. 377--389, 2019.

\bibitem{abbaspour2022online}
M.~Abbaspour and M.~A. Masnadi-Shirazi, ``Online multi-object tracking with
  $\delta$-glmb filter based on occlusion and identity switch handling,''
  \emph{Image and Vision Computing}, vol. 127, p. 104553, 2022.

\bibitem{wojke2017simple}
N.~Wojke, A.~Bewley, and D.~Paulus, ``Simple online and realtime tracking with
  a deep association metric,'' in \emph{2017 IEEE international conference on
  image processing (ICIP)}.\hskip 1em plus 0.5em minus 0.4em\relax IEEE, 2017,
  pp. 3645--3649.

\bibitem{kuhn1955hungarian}
H.~W. Kuhn, ``The hungarian method for the assignment problem,'' \emph{Naval
  research logistics quarterly}, vol.~2, no. 1-2, pp. 83--97, 1955.

\bibitem{jonker1988shortest}
R.~Jonker and T.~Volgenant, ``A shortest augmenting path algorithm for dense
  and sparse linear assignment problems,'' in \emph{DGOR/NSOR: Papers of the
  16th Annual Meeting of DGOR in Cooperation with NSOR/Vortr{\"a}ge der 16.
  Jahrestagung der DGOR zusammen mit der NSOR}.\hskip 1em plus 0.5em minus
  0.4em\relax Springer, 1988, pp. 622--622.

\bibitem{zhang2021fairmot}
Y.~Zhang, C.~Wang, X.~Wang, W.~Zeng, and W.~Liu, ``{FairMOT}: On the fairness
  of detection and re-identification in multiple object tracking,''
  \emph{International Journal of Computer Vision}, vol. 129, pp. 3069--3087,
  2021.

\bibitem{ge2021yolox}
Z.~Ge, S.~Liu, F.~Wang, Z.~Li, and J.~Sun, ``Yolox: Exceeding yolo series in
  2021,'' \emph{arXiv preprint arXiv:2107.08430}, 2021.

\bibitem{redmon2016you}
J.~Redmon, S.~Divvala, R.~Girshick, and A.~Farhadi, ``You only look once:
  Unified, real-time object detection,'' in \emph{Proceedings of the IEEE
  conference on computer vision and pattern recognition}, 2016, pp. 779--788.

\bibitem{wang2020towards}
Z.~Wang, L.~Zheng, Y.~Liu, Y.~Li, and S.~Wang, ``Towards real-time multi-object
  tracking,'' in \emph{European Conference on Computer Vision}.\hskip 1em plus
  0.5em minus 0.4em\relax Springer, 2020, pp. 107--122.

\bibitem{zhang2022bytetrack}
Y.~Zhang, P.~Sun, Y.~Jiang, D.~Yu, Z.~Yuan, P.~Luo, W.~Liu, and X.~Wang,
  ``{ByteTrack}: Multi-object tracking by associating every detection box,'' in
  \emph{European Conference on Computer Vision}.\hskip 1em plus 0.5em minus
  0.4em\relax Springer, 2022, pp. 1--21.

\bibitem{milan2016mot16}
A.~Milan, L.~Leal-Taix{\'e}, I.~Reid, S.~Roth, and K.~Schindler, ``{MOT16}: A
  benchmark for multi-object tracking,'' \emph{arXiv preprint
  arXiv:1603.00831}, 2016.

\bibitem{el2022multiple}
W.~A. El~Ahmar, D.~Kolhatkar, F.~E. Nowruzi, H.~AlGhamdi, J.~Hou, and
  R.~Laganiere, ``Multiple object detection and tracking in the thermal
  spectrum,'' in \emph{Proceedings of the IEEE/CVF Conference on Computer
  Vision and Pattern Recognition}, 2022, pp. 277--285.

\bibitem{stalder2010cascaded}
S.~Stalder, H.~Grabner, and L.~Van~Gool, ``Cascaded confidence filtering for
  improved tracking-by-detection,'' in \emph{Computer Vision--ECCV 2010: 11th
  European Conference on Computer Vision, Heraklion, Crete, Greece, September
  5-11, 2010, Proceedings, Part I 11}.\hskip 1em plus 0.5em minus 0.4em\relax
  Springer, 2010, pp. 369--382.

\bibitem{bewley2016simple}
A.~Bewley, Z.~Ge, L.~Ott, F.~T. Ramos, and B.~Upcroft, ``Simple online and
  realtime tracking,'' in \emph{2016 IEEE international conference on image
  processing (ICIP)}.\hskip 1em plus 0.5em minus 0.4em\relax IEEE, 2016, pp.
  3464--3468.

\bibitem{kalman1960contributions}
R.~E. Kalman \emph{et~al.}, ``Contributions to the theory of optimal control,''
  \emph{Bol. soc. mat. mexicana}, vol.~5, no.~2, pp. 102--119, 1960.

\bibitem{bochinski2017high}
E.~Bochinski, V.~Eiselein, and T.~Sikora, ``High-speed tracking-by-detection
  without using image information,'' in \emph{2017 14th IEEE international
  conference on advanced video and signal based surveillance (AVSS)}.\hskip 1em
  plus 0.5em minus 0.4em\relax IEEE, 2017, pp. 1--6.

\bibitem{blackman1999design}
S.~Blackman and R.~Popoli, ``Design and analysis of modern tracking
  systems(book),'' \emph{Norwood, MA: Artech House, 1999.}, 1999.

\bibitem{bochinski2018extending}
E.~Bochinski, T.~Senst, and T.~Sikora, ``Extending {IOU} based multi-object
  tracking by visual information,'' in \emph{2018 15th IEEE International
  Conference on Advanced Video and Signal Based Surveillance (AVSS)}.\hskip 1em
  plus 0.5em minus 0.4em\relax IEEE, 2018, pp. 1--6.

\bibitem{lyu2017ua}
S.~Lyu, M.-C. Chang, D.~Du, L.~Wen, H.~Qi, Y.~Li, Y.~Wei, L.~Ke, T.~Hu,
  M.~Del~Coco \emph{et~al.}, ``{UA-DETRAC} 2017: Report of avss2017 \& iwt4s
  challenge on advanced traffic monitoring,'' in \emph{Advanced Video and
  Signal Based Surveillance (AVSS), 2017 14th IEEE International Conference
  on}.\hskip 1em plus 0.5em minus 0.4em\relax IEEE, 2017, pp. 1--7.

\bibitem{chen2018real}
L.~Chen, H.~Ai, Z.~Zhuang, and C.~Shang, ``Real-time multiple people tracking
  with deeply learned candidate selection and person re-identification,'' in
  \emph{2018 IEEE international conference on multimedia and expo
  (ICME)}.\hskip 1em plus 0.5em minus 0.4em\relax IEEE, 2018, pp. 1--6.

\bibitem{wang2021joint}
Y.~Wang, K.~Kitani, and X.~Weng, ``Joint object detection and multi-object
  tracking with graph neural networks,'' in \emph{2021 IEEE International
  Conference on Robotics and Automation (ICRA)}.\hskip 1em plus 0.5em minus
  0.4em\relax IEEE, 2021, pp. 13\,708--13\,715.

\bibitem{kim2015multiple}
C.~Kim, F.~Li, A.~Ciptadi, and J.~M. Rehg, ``Multiple hypothesis tracking
  revisited,'' in \emph{Proceedings of the IEEE international conference on
  computer vision}, 2015, pp. 4696--4704.

\bibitem{musicki2004joint}
D.~Musicki and R.~Evans, ``Joint integrated probabilistic data association:
  Jipda,'' \emph{IEEE transactions on Aerospace and Electronic Systems},
  vol.~40, no.~3, pp. 1093--1099, 2004.

\bibitem{bar2005tracking}
Y.~Bar-Shalom, T.~Kirubarajan, and C.~Gokberk, ``Tracking with
  classification-aided multiframe data association,'' \emph{IEEE Transactions
  on Aerospace and Electronic systems}, vol.~41, no.~3, pp. 868--878, 2005.

\bibitem{meza2010steepest}
J.~C. Meza, ``Steepest descent,'' \emph{Wiley Interdisciplinary Reviews:
  Computational Statistics}, vol.~2, no.~6, pp. 719--722, 2010.

\bibitem{dendorfer2020mot20}
P.~Dendorfer, H.~Rezatofighi, A.~Milan, J.~Shi, D.~Cremers, I.~Reid, S.~Roth,
  K.~Schindler, and L.~Leal-Taix{\'e}, ``{MOT20}: A benchmark for multi object
  tracking in crowded scenes,'' \emph{arXiv preprint arXiv:2003.09003}, 2020.

\bibitem{luiten2021hota}
J.~Luiten, A.~Osep, P.~Dendorfer, P.~Torr, A.~Geiger, L.~Leal-Taix{\'e}, and
  B.~Leibe, ``{HOTA}: A higher order metric for evaluating multi-object
  tracking,'' \emph{International journal of computer vision}, vol. 129, pp.
  548--578, 2021.

\bibitem{ristani2016performance}
E.~Ristani, F.~Solera, R.~Zou, R.~Cucchiara, and C.~Tomasi, ``Performance
  measures and a data set for multi-target, multi-camera tracking,'' in
  \emph{European conference on computer vision}.\hskip 1em plus 0.5em minus
  0.4em\relax Springer, 2016, pp. 17--35.

\bibitem{bernardin2008evaluating}
K.~Bernardin and R.~Stiefelhagen, ``Evaluating multiple object tracking
  performance: the clear mot metrics,'' \emph{EURASIP Journal on Image and
  Video Processing}, vol. 2008, pp. 1--10, 2008.

\bibitem{dendorfer2021motchallenge}
P.~Dendorfer, A.~Osep, A.~Milan, K.~Schindler, D.~Cremers, I.~Reid, S.~Roth,
  and L.~Leal-Taix{\'e}, ``Motchallenge: A benchmark for single-camera multiple
  target tracking,'' \emph{International Journal of Computer Vision}, vol. 129,
  pp. 845--881, 2021.

\end{thebibliography}

\end{document}